\documentclass[sigconf]{acmart}
\AtBeginDocument{%
  \providecommand\BibTeX{{%
    \normalfont B\kern-0.5em{\scshape i\kern-0.25em b}\kern-0.8em\TeX}}}

\copyrightyear{2023}
\acmYear{2023}
\setcopyright{acmlicensed}
\acmConference[WWW '23]{Proceedings of the ACM
Web Conference 2023}{April 30-May 4, 2023}{Austin, TX, USA}
\acmBooktitle{Proceedings of the ACM Web Conference 2023 (WWW '23),  April
30-May 4, 2023, Austin, TX, USA}
\acmPrice{15.00}
\acmDOI{10.1145/3543507.3583549}
\acmISBN{978-1-4503-9416-1/23/04}

\usepackage{footnote}
\usepackage{lineno}
\usepackage{algorithmic}
\usepackage{subfigure}
\usepackage{textcomp}
\usepackage[utf8]{inputenc} 
\usepackage[T1]{fontenc}    
\usepackage{hyperref}       
\usepackage{url}           
\usepackage{booktabs}       
\usepackage{amsmath,amsfonts,amssymb,graphicx} 
\usepackage{nicefrac}       
\usepackage{microtype}      
\usepackage{xcolor}  
\usepackage{wrapfig} 
\usepackage{enumitem}
\usepackage{multirow}

\begin{document}
\title{KGTrust: Evaluating Trustworthiness of SIoT via Knowledge Enhanced Graph Neural Networks}

\author{Zhizhi Yu$^{1}$, Di Jin$^{1}$, Cuiying Huo$^{1}$, Zhiqiang Wang$^{1}$, Xiulong Liu$^{1}$, Heng Qi$^{2}$, Jia Wu$^{3}$, Lingfei Wu$^{4}$}
\affiliation{
\institution{$^1$College of Intelligence and Computing, Tianjin University, Tianjin, China} 
\institution{$^2$School of Computer Science and Technology, Dalian University of Technology, Liaoning, China}
\institution{$^3$School of Computing, Macquarie University, Sydney, Australia}
\institution{$^4$Content and Knowledge Graph, Pinterest, New York, USA}
\country{}
}

% \renewcommand{\shortauthors}{}

\begin{abstract}
Social Internet of Things (SIoT), a promising and emerging paradigm that injects the notion of social networking into smart objects (i.e., things), paving the way for the next generation of Internet of Things. 
However, due to the risks and uncertainty, a crucial and urgent problem to be settled is establishing reliable relationships within SIoT, that is, trust evaluation.
Graph neural networks for trust evaluation typically adopt a straightforward way such as one-hot or node2vec to comprehend node characteristics, which ignores the valuable semantic knowledge attached to nodes. Moreover, the underlying structure of SIoT is usually complex, including both the heterogeneous graph structure and pairwise trust relationships, which renders hard to preserve the properties of SIoT trust during information propagation. To address 
these aforementioned problems, we propose a novel knowledge-enhanced graph neural network (KGTrust) for better trust evaluation in SIoT. Specifically, we first extract useful knowledge from users’ comment behaviors and external structured triples related to object descriptions, in order to gain a deeper insight into the semantics of users and objects. Furthermore, we introduce a discriminative convolutional layer that utilizes heterogeneous graph structure, node semantics, and augmented trust relationships to learn node embeddings from the perspective of a user as a trustor or a trustee, effectively capturing multi-aspect properties of SIoT trust during information propagation. Finally, a trust prediction layer is developed to estimate the trust relationships between pairwise nodes. Extensive experiments on three public datasets illustrate the superior performance of KGTrust over state-of-the-art methods.
\end{abstract}

\begin{CCSXML}
<ccs2012>
<concept>
<concept_id>10002950.10003624</concept_id>
<concept_desc>Mathematics of computing~Graph algorithms</concept_desc>
<concept_significance>500</concept_significance>
</concept>
\ccsdesc[500]{Mathematics of computing~Discrete mathematics}
<concept>
<concept_id>10010147.10010257.10010293.10010294</concept_id>
<concept_desc>Computing methodologies~Neural networks</concept_desc>
<concept_significance>500</concept_significance>
</concept>
</ccs2012>
\end{CCSXML}

\ccsdesc[500]{Mathematics of computing~Graph algorithms}
\ccsdesc[300]{Computing methodologies~Neural networks}

\keywords{Graph Neural Networks, Trust Evaluation, Social Internet of Things}


\maketitle
\section{Introduction}
With the rapid development of Internet of Things (IoT) and communication technology, Social Internet of Things (SIoT) \cite{DBLP:journals/icl/AtzoriIM11, DBLP:conf/infocom/XiaXZHC19}, which integrates the concept of social networking into the IoT ecosystem, has aroused considerable research
interest. 
The typical paradigm of SIoT involves a massive number of users, objects (or things) and their associations (user-user, user-object, as well as object-object).
However, due to the inherent openness of SIoT, it is inevitable that malicious users or objects spread
incorrect information or launch illegal attacks, inflicting serious damage to the availability and integrity of network resources.
In addition, as users increasingly depend on
intelligent and interconnected objects in every aspect of life, the demand for security and privacy is growing significantly.
Therefore, it is of great importance and urgent to effectively establish reliable relationships within SIoT.

Trust evaluation \cite{DBLP:journals/csur/WangJYFPY20, DBLP:journals/tkdd/GaoXLC21}, which aims to evaluate the unobserved trust relationship between a pair of nodes, is considered an effective method for assessing SIoT' credibility and reliability. 
To date, an extensive amount of algorithms for trust evaluation have been proposed. 
For example, several studies adopt the idea of random walk \cite{DBLP:conf/infocom/LiuCYZWW17, DBLP:journals/tsc/LiuWOL13}, employing trust propagation along the path from the source node to the target node to assess trustworthiness. 
Another line of attempt is the matrix factorization-based method \cite{DBLP:conf/wsdm/TangGHL13, DBLP:conf/www/YaoTYXL13}, which factorizes a trust matrix into low-rank
embeddings of nodes and their correlations by incorporating prior
knowledge or node-related information. However, these methods either rely heavily on the observed trust relationships or incur the high computational overhead, limiting the performance of trust evaluation. 

Recently, graph neural networks (GNNs) \cite{GNNBook2022,  DBLP:journals/tkde/JinYJPHWYZ23,  DBLP:conf/icdm/YuJLHWT021},which exhibit significant power in
naturally capturing both graph structure and node features, have gained great success and paved a new way for trust evaluation. 
For instance, Guardian \cite{DBLP:conf/infocom/LinGL20} estimates the trustworthiness among two nodes by designing a GNN-based model that captures both graph structure and associated trust relationships. 
GATrust \cite{tkde-gatrust} presents a GNN-driven approach which integrates multiple node attributes and observed trust interactions to predict the trustworthiness of pairwise nodes. 
However, these methods suffer from an inability to well
comprehend and analyze node semantics, as they usually initialize node embeddings via straightforward ways such as one-hot or node2vec.
More importantly, the observed trust relationships in real-world are often very sparse \cite{DBLP:series/synthesis/2015Tang2}, which makes it difficult for these methods to fully model the multi-aspect properties of SIoT trust during information propagation, thus negatively influencing the prediction of trust relationships.

So an interesting yet important question is how to effectively design a GNN-based method for more accurate
% in order to make an accurate 
trust evaluation within SIoT.
Particularly, two challenges need to be addressed. 
First, the rich node semantics within SIoT should be taken into consideration.
As nodes in SIoT are generally associated with textual information such as comments or descriptions,
it is crucial to deeply mine and encode these useful data to essentially embody the inherent characteristics of nodes.
Second, multi-aspect properties of SIoT trust should be effectively preserved. 
The underlying structure of SIoT is usually complex, which contains not only the heterogeneous graph structure but also the pairwise trust relationships. 
Therefore, a reliable GNN-based method ought to make allowance for
flexibly preserving the properties of SIoT trust (including asymmetric, propagative, and composable nature) when information propagates along the heterogeneous graph structure, especially when the observed trust connections are very sparse.

In light of the aforementioned challenges, we propose KGTrust, a knowledge enhanced graph neural network model for trust evaluation in SIoT. Specifically, we first design an embedding layer to fully model the semantics of users and objects by extracting useful and relevant knowledge from users’ comment behaviors and external structured triples, respectively.
We then introduce a personalized PageRank-based neighbor sampling strategy to augment the trust structure, alleviating the sparsity of user-specific trust relationships.
After that, we employ a discriminative convolutional mechanism to learn node embeddings from the perspective of a user as a trustor or a trustee, and adaptively integrate them with a learnable gating mechanism. 
In this way, multi-aspect properties of SIoT trust, including asymmetric, propagative and composable nature, can be effectively preserved. 
Finally, the learned embeddings for pairwise users are concatenated and fed into a prediction layer to estimate their trust relationships.

We summarize our main contributions as follows:
\begin{itemize}
\item To the best of our knowledge, we are the first to gain a deeper insight into the trust evaluation within SIoT via jointly considering three key ingredients, that is, 
% multi-view node information 
heterogeneous graph
structure, node semantics and associated trust relationships.

\item We present a novel knowledge enhanced graph neural network, named KGTrust, which innovatively mines and models the intrinsic characteristics of users and objects with the guidance of external knowledge and multi-aspect trust properties, for assessing trustworthiness in SIoT.

\item Extensive experiments across three public datasets demonstrate the superior performance of the new approach KGTrust over state-of-the-art baselines.
\end{itemize}

% The rest of this paper is organized as follows. Section \ref{section2} gives the preliminaries. Section \ref{section3} describes our proposed knowledge enhanced graph neural networks for trust evaluation.
% We conduct experiments in Section \ref{section4}. Finally, we discuss related work in Section \ref{section5} and conclude in Section \ref{section6}.

\section{Preliminaries}
\label{section2}
We first give the notations and problem definition, then introduce properties of SIoT trust, and finally discuss graph neural networks as the base of our proposed KGTrust.

\subsection{Notations and Problem Definition}
\textbf{Definition 1. Social Trust Internet of Things.} A Social Trust Internet of Things, defined as $G = (V, E, \mathcal{A}, \mathcal{R}, \psi, \varphi)$, is a form of heterogeneous directed network, where $V=\left\{v_{1}, \dots, v_{n}\right\}$ and $E = \{e_{ij}\} \subseteq V \times V$
represent the sets of nodes and edges, respectively.
% Nodes $V = (V_U, V_O)$ are divided into user nodes $V_U$ and object nodes
% $V_O$, where user nodes have associated textual descriptions of their interested
% objects, and object nodes have attached object name.
It is also associated with a node type mapping function $\psi: {V} \rightarrow \mathcal{A}$, and an edge type mapping function $\varphi: {E} \rightarrow \mathcal{R}$, where $\mathcal{A} \in \{$user, object$\}$ and $\mathcal{R} \in$ \{$\langle\rm user, user \rm\rangle$, $\langle\rm user, object \rm\rangle$, 
$\langle\rm object, user \rm\rangle$, $\langle\rm object, object \rm\rangle$\} denote the sets of node and edge types. 
All edges formulate an original adjacency matrix $\bold{A} = (a_{ij})_{n\times n}$, where $a_{ij}$ denotes the relation between nodes $v_i$ and $v_j$. Notice that, the edges representing the trust relationships between user nodes are asymmetric while the others are symmetric.

\textbf{Definition 2. Trust Evaluation.} Given a Social Trust Internet of Things $G$,
let $T = \{\langle v_i, v_j\rangle, t_{ij}|e_{ij} \in E\}$ be the set of observed trust relationships of user nodes, where nodes $v_i$ and $v_j$ denote trustor and trustee, respectively, and $t_{ij}$ measures the trustworthiness from nodes $v_i$ to $v_j$, which is typically application specific. For example, in {SIGCOMM-2009 dataset\footnote{https://crawdad.org/thlab/sigcomm2009}}, trustworthiness is simply divided into two types, that is, trust or distrust.
Trust evaluation is to design a mapping $\mathcal{F}$ to evaluate the trustworthiness of unobserved/missing trust relationship of the trustor-trustee pair $\overline{t}_{ij}$, where $v_i, v_j \in V$, $v_i \neq v_j$, and $e_{ij} \notin E$.
Frequently used
notations are summarized in Appendix A.
% The notations we used throughout the paper are summarized in Table \ref{notations}.

\subsection{Properties of SIoT Trust}
To essentially establish the trustworthiness between pairwise user nodes within SIoT, multi-aspect trust properties, including asymmetric, propagative and aggregative, should be considered. 
% taken into account.

\begin{figure}[htp]
\vspace{-0.5cm}
\setlength{\abovecaptionskip}{0.2cm}
\centering
\subfigure[Asymmetric Nature]{
% \begin{minipage}[t]{0.2\textwidth}
\includegraphics[width=0.475\linewidth]{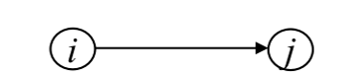}
% \end{minipage}
}
\subfigure[Propagative Nature]{
% \begin{minipage}[t]{0.2\linewidth}
\includegraphics[width=0.475\linewidth]{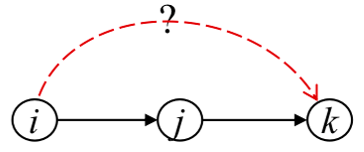}
% \end{minipage}
}\\
\subfigure[Composable Nature]{
% \begin{minipage}[t]{0.2\linewidth}
\includegraphics[width=0.475\linewidth]{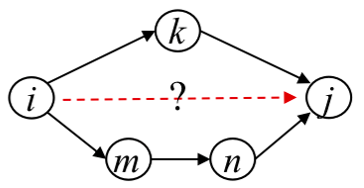}
% \end{minipage}
}
\caption{The  illustration of properties of SIoT trust. \label{Chara_Trust}}
\vspace{-0.5cm}
\end{figure}

\textbf{Asymmetric Nature.} 
The trust relationship between nodes is unequal, that is, node $v_i$ trusts node $v_j$ does not mean node $v_j$ trusts node $v_i$, shown as Figure \ref{Chara_Trust}(a).
Formally, let $t_{ij}$ be the trustworthiness of trustor-trustee pair $\langle v_i, v_j\rangle$, the asymmetry nature of trust is expressed as:
\begin{equation}
t_{ij} \neq t_{ji}.
\end{equation}

\textbf{Propagative Nature.} It indicates that trust may be propagated from one node to another, creating a trust chain for two nodes that are not explicitly connected. 
As shown in Figure \ref{Chara_Trust}(b), assuming that node $v_i$ trusts node $v_j$ and node $v_j$ trusts node $v_k$, 
\begin{figure*}
    \centering
    \includegraphics[width=0.95\linewidth]{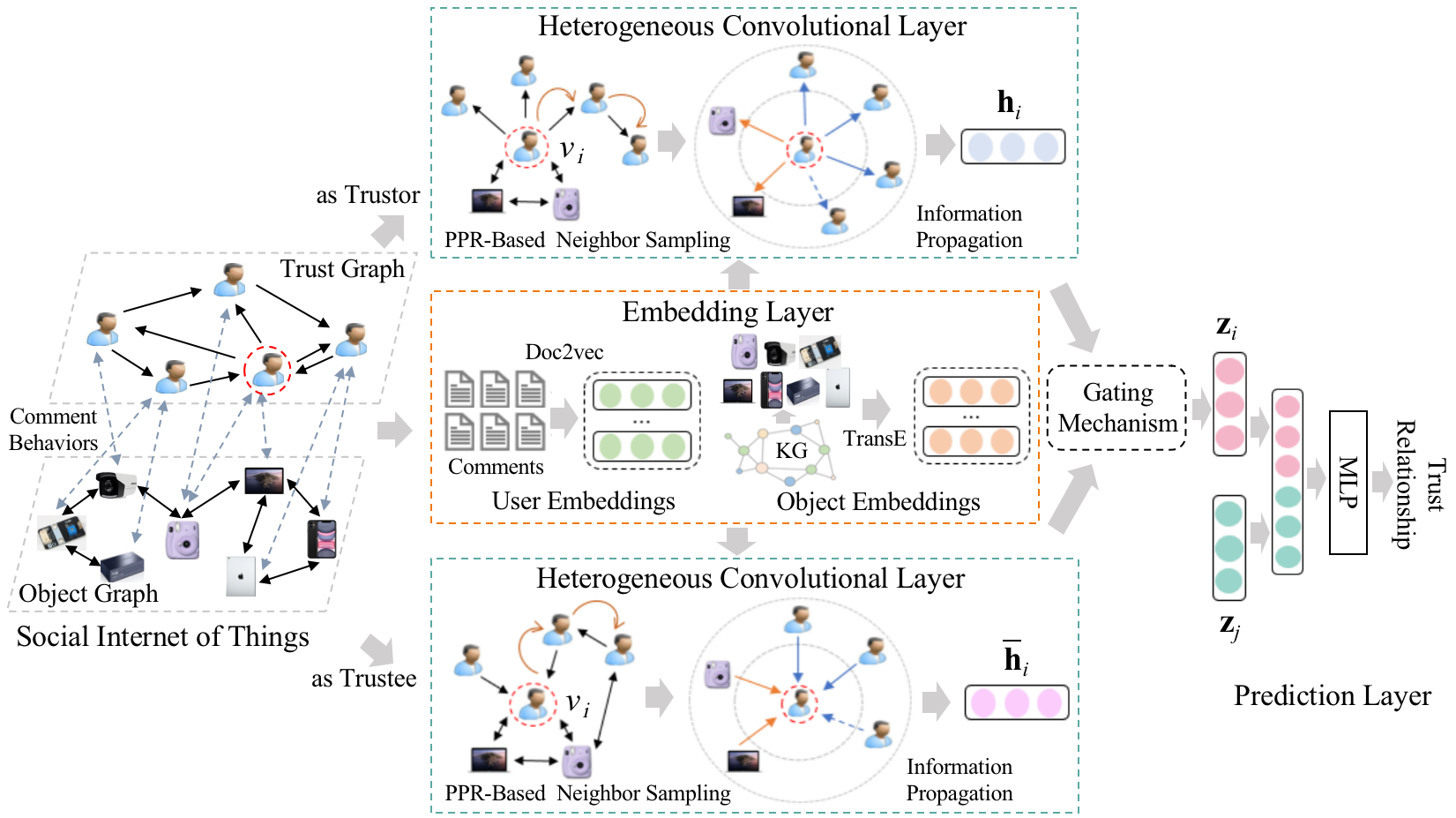}
    \caption{The architecture of KGTrust, which is constituted of three key components: 1) \textbf{Embedding Layer}: a comprehensive user and object modeling by integrating user comments and external knowledge triples; 2) \textbf{Heterogeneous Convolutional Layer}:
    a knowledge enhanced
    graph neural network to further mine and learn node latent embeddings; as well as 3) \textbf{Prediction Layer}: measuring the trust relationships between user pairs.
    }
    \label{KGTrust}
\vspace{-0.3cm}
\end{figure*}
it can then be inferred
that node $v_i$ may trust node $v_k$ to a certain extent. The propagation nature of trust is defined as: 
\begin{equation}
t_{ij} \wedge t_{jk} \Rightarrow{t_{ik}}.
\end{equation}

\textbf{Composable Nature.} It refers to the fact that the trustworthiness propagated by a node from different trust chains to another non-neighbor node can be aggregated.
% where the simplest way is to sum first and then average. 
For example, in Figure \ref{Chara_Trust}(c), there are two trust chains among node $v_i$ and node $v_j$, that is, $v_i \rightarrow{v_k} \rightarrow{v_j}$ and $v_i \rightarrow{v_m} \rightarrow{v_n} \rightarrow{v_j}$. As a result, it is necessary to aggregate the trustworthiness from these two trust chains to determine the trust relationship 
from node $v_i$ to node $v_j$.

\subsection{Graph Neural Networks}
Graph neural networks (GNNs) are a kind of neural networks that directly operate on graph-structured data \cite{ DBLP:conf/iclr/VelickovicCCRLB18, chen2020iterative}. They typically follow the message passing framework,  which learns embedding of each node through iteratively
propagating and aggregating feature information from its topological
neighbors. 
Mathematically, let $\bold{h}_i^{(l)}$ be the latent embedding of node $v_i$ at the $l$-th layer, the message passing process is defined as:
\begin{equation}
\begin{split}
&\bold{m}_i^{(l)}= \operatorname{AGG}^{(l)}(\{\bold{h}_j^{(l-1)}: v_j \in \mathcal{N}(v_i)\}), \quad \\
&\bold{h}_i^{(l)} = \operatorname{UPD}^{(l)}(\bold{h}_i^{(l-1)}, \bold{m}_i^{(l)}), 
\end{split}
\end{equation}
where AGG and UPD denote the functions to aggregate and update the message, $\bold{h}_i^{(0)}$ represents the node’s attributes, and $\mathcal{N}(v_i)$ denotes the set of neighbors of node $v_i$.

% More specifically, GCN \cite{DBLP:conf/iclr/KipfW17}, one of the most representative studies of graph neural networks, is defined as:
% \begin{equation}
% \begin{split}
% \bold{H}^{(l)} =  \operatorname{\sigma}( \bold{\hat{A}} \cdot \bold{H}^{(l-1)} \cdot \bold{W}^{(l-1)}),
% \end{split}
% \end{equation}
% where $\bold{\hat{A}}=\bold{\tilde{D}}^{-\frac{1}{2}}\bold{\tilde {A}} \bold{\tilde {D}}^{-\frac{1}{2}}$ (${\bold{\tilde {A}} = \bold{A}+\bold{I}}$ stands for the adjacency matrix with self-loops, and $\bold{\tilde D}$ = diag $(\tilde d_1,...,\tilde d_n)$ with $\tilde d_i = \sum_j{\tilde a_{ij} }$) is the normalized adjacency matrix, $\bold{W}^{(l-1)}$ is the layer-specific trainable transformation matrix, and $\sigma$ is the non-linear activation function such as ReLU.

\section{METHODOLOG}
\label{section3}
We first give a brief overview of the proposed method, and then introduce three key components in detail. 

\subsection{Overview}
To effectively assess the potential trust relationships within SIoT, we propose a novel knowledge enhanced graph neural network that can fully mine the inherent characteristics of nodes and model the multi-aspect properties of SIoT trust, namely KGTrust. The whole structure of KGTrust is illustrated in Figure \ref{KGTrust}, which consists of three main components, that is, embedding layer, heterogeneous convolutional layer as well as predictor layer.
Specifically, for the embedding layer, we initialize the embeddings of users and objects by extracting useful and related knowledge from
users' comment behaviors and external structured triples, so as to fully explore the intrinsic characteristics of users and objects. 
For the heterogeneous convolutional layer, we utilize the propagative nature of SIoT trust to augment the trust structure, and then introduce a discriminative convolutional mechanism to learn both node and object embeddings by considering the role of a user as a trustor or a trustee, respectively.
After that, we leverage a learnable gating mechanism to adaptively integrate these two types of user embeddings, capturing the asymmetric nature of SIoT trust.
For the prediction layer, a single-layer multiple perceptron is introduced to predict the trust relationships between user pairs based on user embeddings.

% we construct a flexible heterogeneous document-entity framework for 
% modeling the text-rich network, which 
% entails both local and global semantic relationships among documents and entities.
% For the knowledge-based entity representation, we extract useful and related knowledge from structured triples and unstructured entity description, and accordingly learn jointly entity representations by automatically finding a balance
% between these two types of information. 
% This provides sufficient information for understanding text semantics and further enhance its integration with network structures.
% For the bidirectional graph attention, we introduce a discriminative propagation mechanism which adaptively aggregates the information from both text-rich network data space and knowledge space, enabling the model to realize a well balanced combination of network structure and textual semantics.

\subsection{Embedding Layer}
To gain a deeper insight into 
% comprehensively mine and utilize 
the inherent characteristics of users and objects within SIoT, we initialize user and object embeddings, which is the cornerstone of GNNs, by taking account of the users' comment behaviors and external knowledge related to object descriptions, respectively.

\textbf{User Embedding.} 
In SIoT, users typically deliver their opinions on objects by providing comments in the form of text, which reflects the characteristics of users to a certain extent. 
Based on this, 
% Therefore, 
for each user, we employ Doc2vec \cite{DBLP:conf/icml/LeM14}, an unsupervised algorithm to learn
fixed-length node embeddings for texts with variable lengths, to initialize its embedding. 
Specifically, for a user node $v_i$, let ${d}_i$ be the set containing all the comments delivered by $v_i$, the user embedding $\bold{h}_i$ can then be calculated as:
\begin{equation}
\bold{h}_i = \operatorname{Doc2vec}(d_i).
\end{equation} 

\textbf{Object Embedding.} For object nodes, we
capture their characteristics by integrating structured knowledge associated with object descriptions  (i.e., head-predicate-tail triplets) from the knowledge graph. 
Here we employ TransE \cite{DBLP:conf/nips/BordesUGWY13}, a simple and effective approach, to parameterize triplets to learn object embeddings. It encodes the head (or tail) node as a low-dimensional embedding and the relation as algebraic operations between head and tail embeddings. 
Given a triplet $(h, r, t)$, let $\bold{r}$ be the embedding of relation $r$, $\bold{h}$ and $\bold{t}$ be the embeddings of objects $h$ and $t$, respectively.
TransE aims to embed each object and relation by optimizing the translation principle $\bold{h} + \bold{r} \approx \bold{t}$, if $(h, r, t)$ holds. 
The score function is formulated as:
\begin{equation}
f(h, r, t) = - ||\bold{h} + \bold{r} - \bold{t}||^{2}_{2}, 
\end{equation}
where $\bold{h}$ and $\bold{t}$ are subject to the normalization constraint that the magnitude of each vector is 1. Intuitively, a large score of $f(h,r,t)$ indicates that the triplet is more likely to be a true fact in real-world, and vice versa. 
Note that we only consider triplets where object nodes within the SIoT are head instead of tail.
In this way, the object embeddings can be effectively enriched at a semantic level.

\textbf{Embedding Transformation.} Considering that the generated user and object embeddings may have unequal dimensions, or even be lied in different embedding spaces, we need to project these two types of embeddings into the same embedding space.
For a node $v_i$ with type $\psi_i$, we project its embeddings into the same latent embedding space using a type-specific linear transformation $\bold{W}_{\psi_i}$:
\begin{equation}
\bold{h}_i^{\prime} = \bold{W}_{\psi_i} \cdot \bold{h}_i,
% \bold{h}_i^{\prime} = \sigma(\bold{W}_{\phi_i} \cdot \bold{h}_i + \bold{b}_{\phi_i}),
\end{equation} 
where $\bold{h}_i$ and $\bold{h}_i^{\prime}$ are the original and projected embedding of node  $v_i$, respectively. 

\subsection{Heterogeneous Convolutional Layer}
After initializing the user and object embeddings, we further design a heterogeneous convolutional layer, which takes the multi-aspect trust properties into consideration, so as to better assess trustworthiness among users in SIoT.
It mainly consists of three modules: personalized PageRank (PPR)-based neighbor sampling, information propagation and information fusion. 

\textbf{PPR-Based Neighbor Sampling.}
Generally, the available user-specified trust relationships within SIoT are often very sparse, that is, a limited number of user pairs with trust relationships are buried in % a disproportionately large number of
a large proportion of user pairs without trust relationships, making trust evaluation an arduous task \cite{DBLP:conf/wsdm/TangGHL13}. 
To this end, we consider employing personalized PageRank, which shows effectiveness in graph neural networks \cite{DBLP:conf/iclr/KlicperaBG19}, to augment the trust structure.

Personalized PageRank \cite{DBLP:conf/www/Haveliwala02} adopts 
a random walk with restart based strategy that uses the propagative nature of SIoT trust to calculate the correlation between nodes. 
It takes the graph structure as input, and computes a ranking score $p_{ij}$ from source node $v_i$ to target node $v_j$, where the larger $p_{ij}$, the more similar these two nodes.
Formally, given a SIoT $G = (V, E)$, let $\bold{\hat{A}}=\bold{\tilde{D}}^{-\frac{1}{2}}\bold{\tilde {A}} \bold{\tilde {D}}^{-\frac{1}{2}}$ be the normalized adjacency matrix, where ${\bold{\tilde {A}} = \bold{A}+\bold{I}}$ stands for the adjacency matrix with self-loops, the PPR
matrix $\bold{P}$ is calculated as:
\begin{equation}
\bold{P} = (1-\lambda)\bold{\hat {A}} \bold{P} +\lambda\bold{I}, 
\end{equation} 
where $\lambda$ is the reset probability. It is worth noting that we use a push iteration method to compute PPR scores according to the existing work \cite{DBLP:conf/kdd/BojchevskiKPKBR20}, which can be approximated effectively even for very large networks \cite{DBLP:conf/icdm/TongFP06}. 

Then, the augmented trust relationships of each user node $v_i$ can be constructed by choosing its top $k$ PPR neighbors: 
\begin{equation}
N_i =\mathop{\arg\max}\limits_{V'\subset V_U, |V'|=k}{\sum\limits_{v_j\in V'}p_{ij}},
\end{equation} 
where $V_U$ represents the set of user nodes in SIoT.
In this way, several long-range but informative trust relationships can be captured, and further promote the modeling of user nodes.

\textbf{Information Propagation.} 
Due to the asymmetry of trust relationships in SIoT, each user may have dual roles as a trustor or trustee. For this purpose, we consider propagating node embeddings over both trustor role and trustee role, so as to extract two specific embeddings in these two roles. 

Specifically, from the perspective of trustor, we learn the embeddings of users and objects through users' augmented outgoing trust relationships, objects' connections, and interactions between user-object pairs.
Mathematically, given a target node $v_i$, as different types of neighbor nodes (user or object) may have different impacts on it, we employ type-level attention \cite{DBLP:conf/emnlp/HuYSJL19} to learn the importance of different types of neighbor nodes. 
Let $\bold{\hat{A}}_O = [\hat{a}_{ij}]$ be the normalized adjacency matrix which is related to the trustor role, $\bold{h}_\psi$ be the embedding of type $\psi$, which is defined as the sum of the neighbor node embedding $\bold{h}_j^{\prime}$ with node $v_j \in \mathcal{N}_i$ under type $\psi$, that is:
\begin{equation}
\bold{h}_\psi = \sum\nolimits_{v_j} \hat{a}_{ij}\bold{h}_j^{\prime}.
\end{equation} 

Based on the target node embedding $\bold{h}_i^{\prime}$ and its corresponding type embedding $\bold{h}_\psi$, the type-level attention weights can then be calculated as:
\begin{equation}
\alpha_\psi =  \operatorname{softmax}_\psi(\operatorname{\sigma}(\eta_\psi^T [\bold{h}_i^{\prime}, \bold{h}_\psi])),
\end{equation}
where $\eta_\psi$ is the attention vector for the type $\psi$, and softmax is adopted to normalize across all the types.

In addition, considering that different neighbor nodes of the same type could also have different importance, we further apply node-level attention \cite{DBLP:conf/iclr/VelickovicCCRLB18} to learn the weights between nodes of the same type. 
Formally, given a target node $v_i$ with type $\psi$, let $v_j$ be its neighbor node with type $\psi'$, the node-level attention weights can then be computed as:
\begin{equation}
\beta_{ij} =  \operatorname{softmax}_{v_j}(\operatorname{\sigma}(\gamma^T \cdot \alpha_{\psi'} [\bold{h}_i^{\prime}, \bold{h}_j^{\prime}])),
\end{equation}
where $\gamma$ is the attention vector, and softmax is applied to normalize across all the neighbor nodes of the target node $v_i$.

By integrating the above process, the matrix form of the layer-wise propagation rule can be defined as follows:
\begin{equation}
\bold{H}^{(l)} =  \operatorname{\sigma}(\sum\nolimits_{\psi \in \mathcal{A}} \bold{B}_{\psi} \cdot \bold{H}_{\psi}^{(l-1)} \cdot \bold{W}_{\psi}^{(l-1)}),
\end{equation}
where $\mathcal{A}$ is the set of node types in SIoT, and $\bold{B}_{\psi} = (\beta_{ij})_{n \times n}$ represents the attention matrix. In this way, the specific information about trustor role can be obtained.

As for the trustee role, we learn the node embeddings via users’ augmented incoming trust relationships, objects’ connections, and interactions between user-object pairs, which can be calculated in the same way as in trustor role. Therefore,  the specific information about trustee role can be captured by generating the node embeddings $\bold{\overline{H}}$. 

\textbf{Information Fusion.}
In order to achieve the optimal combination of embeddings of a user in different roles (trustee or trustor) for downstream trust evaluation, we introduce a learnable gating mechanism \cite{DBLP:conf/ijcai/XuQCH17} to determine how much the joint embedding depends upon the role of trustor or trustee. Given a user node $v_i$, let $\bold{h}_i$ and $\bold{\overline{h}}_i$ represent its embeddings as trustor role and trustee role, respectively, the joint representation $\bold{z}_i$ can be calculated as:
\begin{equation}
\bold{z}_i = \bold{g}_e \odot \bold{h}_i + (1-\bold{g}_e) \odot \bold{\overline{h}}_i,
\end{equation} 
where $\bold{g}_e$ is a gating vector with elements in $[0,1]$ to balance embeddings, and $\odot$ represents element-wise multiplication. 
Obviously, the joint embedding with gate closer to 0 tends to use the embedding of a user as a trustee; whereas the joint embedding with gate closer to 1 utilizes the embedding of a user as a trustor. More importantly, to constrain the value of each element in $[0, 1]$, we apply sigmoid function to calculate the gate $\bold{g}_e$ as:
\begin{equation}
\bold{g}_e = \operatorname{sigmoid}(\bold{\tilde{g}}_e),
\end{equation} 
where $\bold{\tilde{g}}_e$ is a real-value vector that is learned during training.

\subsection{Predictor Layer}
To convert the learned user embeddings into the latent factor of trust relationship in SIoT, 
for a given user pair $\langle v_i, v_j\rangle$, 
we first concatenate the embeddings of nodes $v_i$ and $v_j$, and then feed them to a multiple perceptron (MLP) followed by a softmax function as: 
\begin{equation}
\tilde{y}_{ij} =  \operatorname{softmax}(\operatorname{MLP}(\bold{z}_i \mathop{\parallel} \bold{z}_j)),
\end{equation} 
where $\mathop{\parallel}$ is the concatenation operator, 
% $\sigma$ denotes the softmax function, 
and $\tilde{y}_{ij}$ is the predicted probability that the user pair $\langle v_i, v_j\rangle$ belongs to a trusted pair or a distrusted pair.

Finally, we define the trust evaluation loss function by using cross entropy as:
\begin{equation}
\mathcal{L}=- \sum\limits_{v_iv_j}{y}_{ij}\operatorname{ln}\tilde{y}_{ij},
\end{equation}
where ${y}_{ij}$ denotes the ground truth of trust relationship of user pair $\langle v_i, v_j\rangle$.
In particular, we employ the back propagation algorithm 
% \cite{1986Learning} 
and Adam optimizer 
% \cite{DBLP:journals/corr/KingmaB14} 
to train the model.

\section{Experiments}
\label{section4}
We first introduce the experimental setup, and then compare the new approach KGTrust with state-of-the-arts in terms of effectiveness and robustness.
We finally present an in-depth analysis of different components of KGTrust and give the parameter analysis. 

\subsection{Experimental Setup}

\textbf{Datasets.} We conduct experiments on three widely used SIoT datasets, namely FilmTrust\footnote{http://www.librec.net/datasets.html}, Ciao\footnote{http://www.cse.msu.edu/\textasciitilde tangjili/trust.html\label{data}} and Epinions\textsuperscript{\ref {data}}, where the basic information is summarized in Table \ref{table:dataset}. 
More details of datasets are provided in Appendix B.1.

\renewcommand\arraystretch{1.3}\begin{table}[htp]
\vspace{-0.8cm}
\setlength{\belowcaptionskip}{0.8cm}
	\centering
	\caption{\label{table:dataset}Statistics of the datasets.}
	\begin{small}
	\setlength{\tabcolsep}{1.5mm}{
	\begin{tabular}{lccc}
	\hline
 & FilmTrust 
	& Ciao & Epinions \\
	\hline
		\#Users & 1508 & 4409 & 8174\\
		\#Objects & 2071 & 12,082 & 11,379\\
      \#Comment Behaviors & - & 136,105 & 306,133\\
        \#Trust Relationships & 1853 & 88,649 & 224,589\\
            Trust Network Density & 0.0008 & 0.0046 & 0.0034\\
		\hline
	\multicolumn{4}{l}{\small “-” denotes no such information provided by the dataset.}
	\end{tabular}
	}
	\end{small}
 	\vspace{-0.2cm}
\end{table}

% \begin{itemize}
% \item \textbf{FilmTrust\footnote{http://www.librec.net/datasets.html}} is extracted from the film review website, which consists of two kinds of information, that is, social trust relationships between users; as well as interactive connections between users and objects.
% % ratings that reflect the users' preference for objects, ranging from 0.5 to 4. 
% For FilmTrust, considering that there is no information about users' comment behaviors and object names, we utilize random vectors to initialize user and object embeddings, which are trainable during the message passing process.

% % and treat them as part of the learnable parameters of neural networks.

% \item \textbf{Ciao\footnote{http://www.cse.msu.edu/\textasciitilde tangjili/trust.html\label{data}}} and \textbf{Epinions}\textsuperscript{\ref {data}} are two ideal who-trusts-whom knowledge-sharing websites that contain four types of information, that is, explicit trust relationships between users; connections between users and objects representing interactive relationships; comment behaviors that reveal users' attitudes and preferences in the form of text; as well as object descriptions that reflect the characteristics of objects.
% For these two datasets, we reserve users with more than 15 comment behaviors and objects with more than 10 comment behaviors.
% \end{itemize}

\renewcommand\arraystretch{1.3}
\begin{table*}[!ht] 
\tabcolsep=0.29cm
\centering
	\caption{\label{table: 90 training} Performance comparions on three SIoT datasets in terms of Accuracy (\%) and F1-Score (\%). (bold: best)}
	\begin{small}
	\begin{tabular}{c|c|cccccccc|c}\hline
	{{Datasets}} & {{Metrics}} & {{GAT}} & {{SGC}} & {{SLF}} & {{STNE}} & {{SNEA}} & {{DeepTrust}} & {{AtNE-Trust}} & {{Guardian}} & {{KGTrust}} \\ \hline
	\multirow{2}*{FilmTrust} & \multirow{1}*{\shortstack{Accuracy}} & 68.29 & 75.61 & 65.55 & 72.87 & 63.91 & 53.05 & 63.11 & 77.74 & \textbf{79.82} \\\cline{2-11} {} & \multirow{1}*{\shortstack{F1-Score}} & 71.74 & 77.14 & 65.65 & 73.27 & 66.67 & 64.63 & 65.13 & 79.78 & \textbf{80.92}\\\hline
	\multirow{2}*{Ciao} & \multirow{1}*{\shortstack{Accuracy}} & 64.28 & 69.93 & 72.17 & 71.33 & 68.97 & 50.17 & 68.23 & 72.17 & \textbf{72.56}\\\cline{2-11}
	        {} & \multirow{1}*{\shortstack{F1-Score}} & 71.36 & 70.34 & 73.39 & 71.38 & 70.83 & 66.52 & 71.50 & 73.50 & \textbf{74.30}\\ \hline
	 \multirow{2}*{Epinions} & \multirow{1}*{\shortstack{Accuracy}} & 72.05 & 78.62 & 80.83 & 79.51 & 74.63 & 58.38 & 74.35 & 80.82 & \textbf{81.39}\\ \cline{2-11}
	        {} & \multirow{1}*{\shortstack{F1-Score}} & 75.57 & 78.76 & 80.95 & 78.57 & 74.92 & 64.80 & 74.88 & 81.11 & \textbf{81.84}
	    \\\hline
	\end{tabular}
	\end{small}
%  	\vspace{-0.2cm}
\end{table*}

\renewcommand\arraystretch{1.3}
\begin{table*}[!ht]
\tabcolsep=0.23cm
	\centering
	\caption{\label{table:training ratio} Performance comparions with different training ratios on three SIoT datasets. (bold: best)}
	\begin{small}
		\scalebox{1.0}{\begin{tabular}{c|c|c|cccccccc|c}
			\hline
				{{Datasets}} &{{Metrics}} & {{Training}} & {{GAT}} & {{SGC}} & {{SLF}} & {{STNE}} & {{SNEA}} & {{DeepTrust}} & {{AtNE-Trust}} & {{Guardian}} & {{KGTrust}} \\ 
			\hline
			\multirow{10}*{FilmTrust} & \multirow{5}*{\shortstack{Accuracy \\(\%)  }} & 50\% & 60.36 & 71.26 & 54.96 & 69.42 & 60.14 & 49.51 & 60.17 & 74.14 & \textbf{74.94}\\
			{} & {} & 60\% & 62.79 & 72.21 & 55.51 & 69.98 & 61.01 & 50.08 & 60.72 & 74.81 & \textbf{76.11}
	        \\ 
	        {} & {} & 70\% & 64.39 & 73.16 & 61.22 & 72.14 & 62.90 & 50.20 & 62.14 & 75.51 & \textbf{78.16}
	        \\ 
	        {} & {} & 80\% & 67.28 & 74.01 & 63.61 & 72.78 & 63.30 & 51.68 & 63.00 & 76.45 & \textbf{79.66}
	        \\ 
	        {} & {} & 90\% & 68.29 & 75.61 & 65.55 & 72.87 & 63.91 & 53.05 & 63.11 & 77.74 & \textbf{79.82}
	        \\ \cline{2-12}
	        {} & \multirow{5}*{\shortstack{F1-Score\\(\%)}} & 50\% & 62.35 & 71.52 & 56.48 & 69.44 & 62.45 & 60.11 & 60.27 & 75.52 & \textbf{75.98}
	        \\
	        {} & {} & 60\% & 63.51 & 72.40 & 57.37 & 70.17 & 62.68 & 60.38  & 61.69 & 76.52 & \textbf{76.73}
	        \\
	        {} & {} & 70\% & 66.80 & 73.83 & 62.23 & 72.17 & 64.83 & 62.20 & 63.42 & 78.56 & \textbf{78.94}
	        \\
	        {} & {} & 80\% & 68.34 & 74.40 & 65.00 & 72.33 & 65.12 & 63.41 & 63.88 & 79.08 & \textbf{80.47}
	        \\
	        {} & {} & 90\% & 71.74 & 77.14 & 65.65 & 73.27 & 66.67 & 64.63 & 65.13 & 79.78 & \textbf{80.92}
	        \\\hline
			\multirow{10}*{Ciao} & \multirow{5}*{\shortstack{Accuracy\\(\%)}} & 50\% & 59.76 & 67.40 & 71.32 & 70.69 & 66.88 & 49.80 & 62.24 & 71.27 & \textbf{71.72}
			\\
			{} & {} & 60\% & 61.03 & 68.29 & 71.66 & 70.87 & 67.82 & 50.01 & 62.66 & 71.62 & \textbf{72.11}
	        \\ 
	        {} & {} & 70\% & 62.17 & 68.39 & 71.89 & 70.92 & 68.15 & 50.03 & 63.52 & 71.90 & \textbf{72.34}
	        \\
	        {} & {} & 80\% & 63.01 & 68.81 & 72.08 & 71.05 & 68.53 & 50.07 & 66.58 & 71.94 & \textbf{72.36}
	        \\ 
	        {} & {} & 90\% & 64.28 & 69.93 & 72.17 & 71.33 & 68.97 & 50.17 & 68.23 & 72.17 & \textbf{72.56}
	        \\ \cline{2-12}
	        {} & \multirow{5}*{\shortstack{F1-Score\\(\%)}} & 50\% & 66.47 & 67.53 & 71.87 & 70.83 & 67.68 & 61.30 & 62.76 & 71.84 & \textbf{72.85}
	        \\
	        {} & {} & 60\% & 68.08 & 68.58 & 72.68 & 70.85 & 68.87 & 61.38 & 63.03 & 72.28 & \textbf{73.11}
	        \\
	        {} & {} & 70\% & 70.61 & 68.78 & 72.88 & 71.07 & 69.45 & 61.77 & 65.37 & 72.67 & \textbf{73.23}
	        \\
	        {} & {} & 80\% & 70.85 & 69.76 & 73.00 & 71.32 & 70.15 & 63.63 & 69.92 & 73.32 & \textbf{74.06}
	        \\
	        {} & {} & 90\% & 71.36 & 70.34 & 73.39 & 71.38 & 70.83 & 66.52 & 71.50 & 73.50 & \textbf{74.30}
	        \\\hline
			\multirow{10}*{Epinions} & \multirow{5}*{\shortstack{ Accuracy\\(\%)}} & 50\% & 61.70 & 77.22 & 79.99 & 79.04 & 73.84 & 55.53 & 71.90 & 80.15 & \textbf{80.59}
			\\
			{} & {} & 60\% & 61.92 & 77.57 & 80.05 & 79.13 & 74.12 & 56.25 & 73.01 & 80.22 & \textbf{80.65}
	        \\ 
	        {} & {} & 70\% & 64.76 & 77.82 & 80.44 & 79.32 & 74.36 & 56.71 & 73.40 & 80.31 & \textbf{80.96}
	        \\ 
	        {} & {} & 80\% & 70.79 & 78.17 & 80.60 & 79.45 & 74.59 & 58.23 & 73.59 & 80.55 & \textbf{81.14}
	        \\ 
	        {} & {} & 90\% & 72.05 & 78.62 & 80.83 & 79.51 & 74.63 & 58.38 & 74.35 & 80.82 & \textbf{81.39}
	        \\ \cline{2-12}
	        {} & \multirow{5}*{\shortstack{F1-Score \\(\%)}} & 50\% & 65.60 & 77.63 & 80.08 & 78.18 & 73.28 & 61.27 & 72.87 & 80.41 & \textbf{81.05}
	        \\
	        {} & {} & 60\% & 66.64 & 77.92 & 80.15 & 78.22 & 73.73 & 63.93 & 73.74 & 80.51 & \textbf{81.11}
	        \\
	        {} & {} & 70\% & 72.67 & 78.05 & 80.46 & 78.46 & 74.19 & 64.10 & 73.80 & 80.58 & \textbf{81.46}
	        \\
	        {} & {} & 80\% & 72.84 & 78.56 & 80.63 & 78.50 & 74.61 & 64.36 & 74.29 & 80.86 & \textbf{81.70}
	        \\
	        {} & {} & 90\% & 75.57 & 78.76 & 80.95 & 78.57 & 74.92 & 64.80 & 74.88 & 81.11 & \textbf{81.84}
	         \\\hline
% 			Heuristic & 170	& 138 & 148	& 50 & 46 & 55 & 118 & 108 & 99\\ \hline
		\end{tabular}}
	\end{small}
 	\vspace{-0.2cm}
\end{table*}

\textbf{Baselines.}
We compare KGTrust with eight state-of-art methods. They include: 1) the network embedding methods GAT \cite{DBLP:conf/iclr/VelickovicCCRLB18},
SGC \cite{DBLP:conf/icml/WuSZFYW19}, SLF \cite{DBLP:conf/kdd/XuH0D19}, STNE \cite{DBLP:conf/icdm/XuHW0DY19} and 
SNEA \cite{DBLP:conf/aaai/LiTZC20}, 
and 2) the trust evaluation methods
DeepTrust \cite{DBLP:conf/icdm/WangZYWHX19}, AtNE-Trust \cite{DBLP:conf/icdm/WangZ00ZX20} and Guardian \cite{DBLP:conf/infocom/LinGL20}. 
More details of baselines are provided in Appendix B.2.

\textbf{Implementation Details.} 
For all baselines, we use the source codes released by their corresponding authors.
For KGTrust, we use Wikipedia anchors to align the mentions extracted from object descriptions of SIoT to Wikidata5M \cite{DBLP:journals/tacl/WangGZZLLT21}, a newly proposed large-scale knowledge graph containing 4M entities and 21M fact triplets. 
We use Adam optimizer with an
initial learning rate of 0.005 and a weight decay of 5e-4.
In addition, we set the layer number of heterogeneous convolutional mechanisms as 2, the reset probability $\lambda$ as 0.15 according to \cite{DBLP:conf/www/ZhangZDS021}, and search the top $k$ for PPR-based neighbor sampling strategy in \{10, 20, 30, 40, 50\}.
As there are only positive links (e.g., observed trust relationships) in our FilmTrust, Epinions and Ciao datasets, a set of unlinked user pairs with the same proportion is randomly selected as the negative instance set for training and testing.  
% The code for KGTrust and the datasets are available at GitHub\footnote{https://anonymous.4open.science/r/KGTrust-1}.
 
We use \textbf{Accuracy} and \textbf{F1-Score}, which are two commonly used metrics in trust evaluation tasks, to measure the performance of our proposed KGTrust and baselines.

% We employ two commonly used metrics, i.e., Accuracy and F1-Score, to evaluate the performance of models.

% We employ two widely used evaluation metrics, i.e. Accuracy and F1-Score, to measure the model's quality. 

\subsection{Performance Comparisons}
We empirically compare KGTrust to baselines from two perspectives, including effectiveness and robustness.

\textbf{Effectiveness.} For different datasets, we set the ratio of training and testing set to 90\%:10\%, and run each method 10 times to report the average results of different models.

As presented in Table \ref{table: 90 training}, we can find that KGTrust performs consistently much better than all baselines across three datasets. 
To be specific, in terms of accuracy,
the improvement of KGTrust over 
different baselines ranges from 2.08\% to 26.77\%, 0.39\% to 22.39\%, and 0.57\% to 23.01\% on
% KGTrust achieves
% up to 2.08\%, 0.39\% and 0.57\% better accurate than the best baseline Guardian on 
FilmTrust, Ciao, and Epinions, respectively. 
In terms of F1-Score,
the improvement of KGTrust over 
different baselines ranges from 1.14\% to 16.29\%, 0.80\% to 7.78\%, and 0.73\% to 17.04\% on
% 1.14\%, 0.80\% and 0.73\% more accurate than the best baseline Guardian on 
these three datasets. 
These results not only demonstrate the superiority of enriching node semantics with node-related knowledge, but also validate the effectiveness of flexibly preserving the multi-aspect properties of SIoT trust during information propagation. 
Particularly, the performance of KGTrust is much better than that of vanilla GAT (i.e., 11.53\%, 8.28\%, 9.34\% relative improvements in accuracy, and 9.18\%, 2.94\%, 6.27\% relative improvements in F1-Score), which further shows the significance of jointly considering
three key ingredients within SIoT, namely heterogeneous graph structure, node semantics and associated trust relationships.
Neither DeepTrust nor AtNE-Trust is so competitive here, which is mainly because they fail to make better use of the information propagation over graph structure, seriously affecting their performance for trust evaluation.

\textbf{Robustness}. To further measure the stable ability of our KGTrust and baselines, we conduct experiments across all training and testing set ratios. The ratio of training set is set as $x$\% and the remaining $(1-x)$\% as the testing set, where $x$ belongs to \{50, 60, 70, 80, 90\} over three datasets. We run each method 10 times and report the average performance in terms of accuracy and F1-Score. 

The results are shown in Table \ref{table:training ratio}. As shown, the proposed method KGTrust always performs the best across different training ratios and datasets.
Specifically, 
% KGTrust achieves the best performance when the training set ratio is 90\%. 
when fewer observed trust relationships are provided, the performances of baselines are surprisingly reduced, especially the classical GNN-based methods such as GAT, while our model still achieves relatively high performance. This demonstrates that our method can better assess trust relationships by validly alleviating data sparsity with personalized PageRank-based neighbor sampling.
Moreover, as the ratio of observed trust relationships increases, KGTrust consistently maintains superior and achieves the best performance when the training set ratio is 90\%, which validates the effectiveness and robustness of the proposed approach.
Also of note, KGTrust outperforms  Guardian, which uses GNN for trust evaluation, in all cases, further indicating the rationality of fully mining the intrinsic characteristics of nodes with the guidance of node-related knowledge.

% on average on all three datasets, KGTrust is 1.0\% and 0.75\% more accurate than the best
% baseline Guardian, which also uses GNN for trust evaluation, in terms of accuracy and F1-Score, respectively, 

\subsection{Ablation Study}
Similar to most deep learning models, our proposed KGTrust also contains some important components that may have a significant impact on the performance.
To test the effectiveness of each component, we conduct experiments by  comparing KGTrust with five variations. 
The variants are as follows: 
1) KGTrust of using random vectors instead of introducing structured triples to initialize object embeddings, named as w/o Triples,
2) KGTrust of removing PPR-based neighbor sampling, named as w/o PPR,
3) KGTrust of removing the trustee role of a node, and aggregating information only from its trustor role, named as w/o Trustee, 
4) KGTrust of removing the trustor role of a node, and aggregating information only from its trustee role, named as w/o Trustor, 
and 5) KGTrust of employing concatenation operator instead of gating mechanism to fuse the two role embeddings (truster or trustee) of a user node, named as KGTrust (Con). 

\begin{table}[h]
\vspace{0cm}
    \small
    \caption{\label{Ablation Study} Comparisons of our KGTrust and its five variants on three SIOT datasets
    in terms of Accuracy (\%) and F1-Score (\%). For FilmTrust, we do not introduce structured triples as no object descriptions  provided, while such information is provided by 
    Ciao and Epinions.}
    \centering
    \renewcommand{\arraystretch}{1.3}
        \scalebox{0.82}{\begin{tabular}{l|cc|cc|cc}
            \hline
     \multirow{2}{*}{Datasets} & \multicolumn{2}{c|}{FilmTrust} & \multicolumn{2}{c|}{Ciao} & \multicolumn{2}{c}{Epinions} \\ \cline{2-7} 
            & Accuracy    & F1-Score      & Accuracy     & F1-Score      & Accuracy    & F1-Score     \\ \hline
            KGTrust & \textbf{79.82} & \textbf{80.92} & \textbf{72.56} & \textbf{74.30} & \textbf{81.39} & \textbf{81.84} \\ 
            - w/o Triples & - & - & 71.10 & 72.48 & 80.51 & 80.86 \\ 
           - w/o PPR & 78.29 & 78.74 & 72.12 & 72.88 & 80.71 & 81.19 \\ 
           - w/o Trustee & 78.13 & 79.18 & 59.07 & 64.60 & 70.73 & 72.10 \\ 
           - w/o Trustor & 77.22 & 78.37 & 60.58 & 65.67 & 70.62 & 72.01  \\ 
            KGTrust (Con) & 76.76 & 77.51 & 59.28 & 64.74 & 70.75 & 73.10 \\ \hline
        \end{tabular}}
\vspace{-0.3cm}
\end{table}

From the results in Table \ref{Ablation Study}, we can draw the following conclusions: 
1) The results of KGTrust are consistently better than its five variants, indicating the effectiveness and necessity of jointly taking into account of heterogeneous graph
structure, node semantics and associated trust relationships within SIoT.
2) Removing structured triples or PPR-based neighbor sampling strategy leads to slight performance drop, which demonstrates the usefulness of employing external knowledge to deeply mine object semantics and  propagative nature of trust to enrich the trust relationships.
3) Neither KGTrust w/o Trustee nor KGTrust w/o Trustor is so competitive here, making us
realize the importance of considering the trust asymmetry nature during information propagation.
4) Compared to KGTrust (Con), the improvement
brought by KGTrust is more significant, which illustrates the rationality of adaptively fusing the embeddings of dual roles of a user.

\subsection{Parameter Analysis}
We investigate the sensitivity of two main parameters, including the top $k$ for PPR-based neighbor sampling and the dimension of final embeddings, on Ciao and 
Epinions datasets. 
Results on FilmTrust dataset can be found in Appendix C.

\begin{figure}[htp]
\vspace{-0.25cm}
\setlength{\abovecaptionskip}{0.2cm}
\centering
\subfigure[Ciao]{
% \begin{minipage}[t]{0.2\linewidth}
\includegraphics[width=0.475\linewidth]{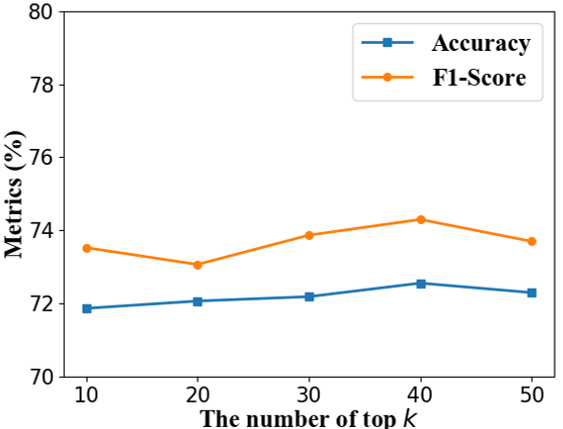}
% \end{minipage}
}
\subfigure[Epinions]{
% \begin{minipage}[t]{0.2\linewidth}
\includegraphics[width=0.475\linewidth]{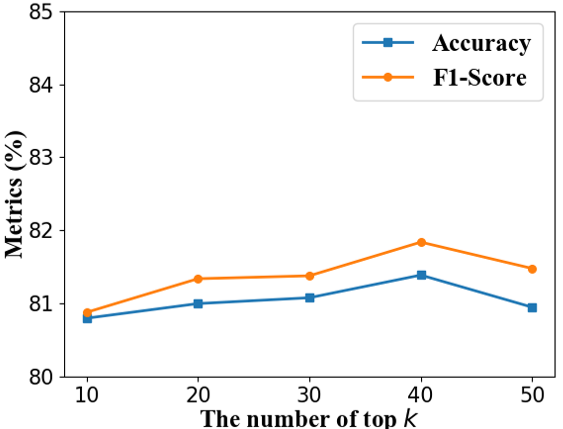}
% \end{minipage}
}
\caption{The performance with different numbers of top $k$ for PPR-based neighbor sampling. \label{Para_k}}
\vspace{-0.25cm}
\end{figure}

\textbf{Analysis of $k$.}
The parameter $k$ determines the number of trust relationships augmented by each user node with PPR-based neighbor sampling.
We vary its value from 10 to 50 and the corresponding results
are shown in Figure \ref{Para_k}. 
With the increase of augmented trust relationships, the performance shows a trend of first rising and then descending.
This is probably because a small number of augmented trust relationships are not enough to obtain informative node embeddings, whereas too many augmented trust relationships may introduce noise and thus weaken the information propagation.

\begin{figure}[htp]
\vspace{-0.2cm}
\setlength{\abovecaptionskip}{0.2cm}
\centering
\subfigure[Ciao]{
% \begin{minipage}[t]{0.2\linewidth}
\includegraphics[width=0.475\linewidth]{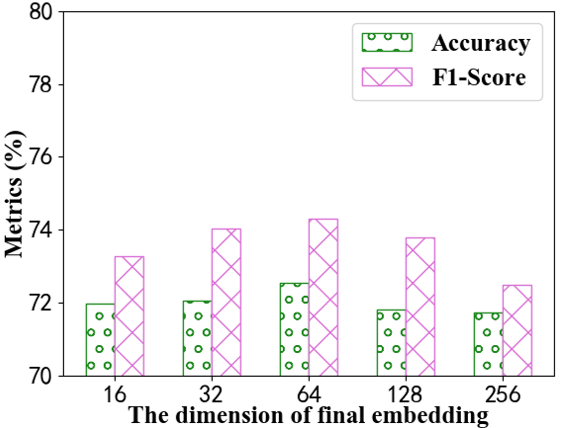}
% \end{minipage}
}
\subfigure[Epinions]{
% \begin{minipage}[t]{0.2\linewidth}
\includegraphics[width=0.475\linewidth]{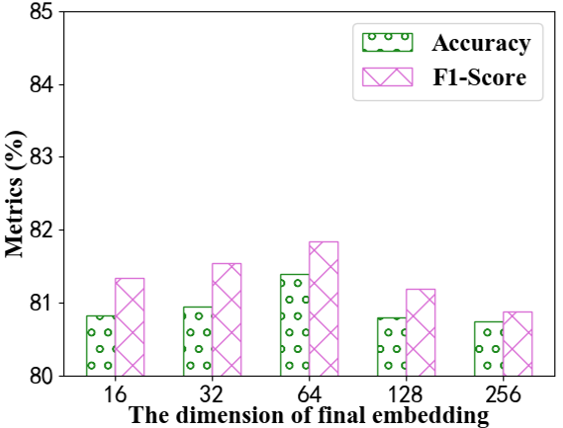}
% \end{minipage}
}
\caption{The performance with different dimensions of final embedding. \label{Para_d}}
\setlength{\belowcaptionskip}{-10cm}
\vspace{-0.4cm}
\end{figure}

\textbf{Analysis of Final Embedding Dimension.} We test the effect of
the dimension of final embedding, and vary it from 16 to 256. 
The result is shown in Figure \ref{Para_d}.
With the increase of the dimension of final embeddings, the values of metrics, including accuracy and F1-Score, increase first and then start to decrease. 
It is reasonable since KGTrust needs a suitable dimension to encode the key ingredients within SIoT, including heterogeneous graph structure, node semantics and associated trust relationships,
while larger dimensions may introduce additional redundancies, affecting the performance of assessing trustworthiness.

\section{RELATED WORK}
\label{section5}
We briefly review some literatures that are related to our work, that is, trust evaluation and knowledge graph neural networks.

% tightly

% In line with the main focus of our work, we review the most related
% literatures in trust evaluation and knowledge graph neural networks.

\subsection{Trust Evaluation}
Trust plays a crucial role in assisting users gather reliable
information and trust evaluation.
predicting the unobserved
pairwise trust relationships among users, draws considerable
research interest. 
Existing trust evaluation methods can be roughly divided into three categories, including random walk-based methods, matrix factorization-based methods and deep learning-based methods.

\textbf{Random Walk-Based methods.}
In the past decade, many attentions have been paid to exploit trust propagation along the path from the trustor to the trustee to assess trustworthiness.
For example, 
ModelTrust \cite{DBLP:conf/aaai/MassaA05} conducts motivating experiments to analyze the difference in accuracy and coverage between local and global trust measures, and then introduces a local trust metric to predict trustworthiness of unknown users. 
OptimalTrust \cite{DBLP:journals/tsc/LiuWOL13} designs a new concept, namely quality of trust, to search the optimal trust path to generate the most trustworthy evaluation.
After that, AssessTrust \cite{DBLP:conf/infocom/LiuYWLW14} introduces three-valued subjective logic (3VSL) that
effectively considers the information in trust propagation and aggregation to assess multi-hop interpersonal trust.
OpinionWalk \cite{DBLP:conf/infocom/LiuCYZWW17} computes the trustworthiness 
between any two users 
based on 3VSL and breadth-first search strategy.

\textbf{Matrix Factorization-Based Methods.} 
Several recent studies have extended matrix factorization to trust evaluation \cite{DBLP:conf/aaai/WangWTZC15}, where the basic idea is to learn the low-rank representations of users and their correlations by incorporating prior knowledge and node-related data.
For instance, 
hTrust \cite{DBLP:conf/wsdm/TangGHL13}
presents an unsupervised method that exploits low-rank matrix factorization and homophily effect for trust evaluation.
mTrust \cite{DBLP:conf/wsdm/TangGL12} argues that the trust relationships among users are usually multiple and heterogeneous, and accordingly designs a fine-grained representation to incorporate multi-faceted trust relationships.
% Furthermore, Matri \cite{DBLP:conf/www/YaoTYXL13} designs a multi-aspect trust inference approach via jointly taking transitivity, multi-aspect and prior knowledge into consideration. 
sTrust \cite{DBLP:conf/aaai/WangWTZC15}
proposes a trust prediction model by 
considering social status theory that reflects users' position or level in the social community.

\textbf{Neural Network-Based Methods.} 
As neural networks become the most eye-catching tools for tracking graphs, several efforts have been devoted to utilizing neural networks to boost the performance of trust evaluation.
For example, 
NeuralWalk \cite{DBLP:conf/infocom/LiuL019} designs a neural network architecture that models single-hop trust propagation and trust combination for assessing trust relationships. 
DeepTrust \cite{DBLP:conf/icdm/WangZYWHX19}
presents a deep trust evaluation model which effectively mines user characteristics by introducing associated user attributes.
AtNE-Trust \cite{DBLP:conf/icdm/WangZ00ZX20} 
improves the performance of trust relationship prediction by jointly capturing the properties of trust network and multi-view user attributes.
C-DeepTrust \cite{C-DeepTrust} points out that the user preference may change due to the drifts of their interests, and accordingly integrates both the static and dynamic user preference to tackle the trust evaluation problem.
Guardian \cite{DBLP:conf/infocom/LinGL20}
estimates the trustworthiness between any two users by simultaneously capturing both
social connections and trust relationships.
GATrust \cite{tkde-gatrust} presents a GNN-driven approach which integrates  multiple node attributes and the observed trust interactions for trust evaluation. 

Despite various trust evaluation algorithms or models have been developed, they still suffer from an inability to comprehensively mine and encode the node semantics within SIoT and multi-aspect properties of SIoT trust.
% which lead to the main contribution in this work: design a knowledge enhanced trust evaluation model based on graph neural network.

% \subsection{Integrating External Knowledge}
% With the advanced development of knowledge base, a large number of large-scale knowledge bases (KBs) such as YAGO and Freebase are available. 

% With the advanced development of knowledge base construction, a large number of large-scale
% knowledge bases (KBs) such as YAGO [42] and Freebase [1] are available. As can be seen from
% many other tasks, it becomes a tendency to exploit external knowledge from KBs to enrich the
% representational learning in deep-learning models

% As can be seen from
% many other tasks, it becomes a tendency to exploit external knowledge from KBs to enrich the representational learning in deep-learning models. Several efforts have been made on integrating knowledge embeddings trained by knowledge embedding methods [54] to learn a knowledgeaware sentence representation on machine reading [60], entity typing [58] and relation extraction [19]. In this article, we integrally model contexts and external knowledge into sentence representations, and present the knowledge-aware attention mechanism to explore the interrelation
% between the knowledge of questions and answers.

\subsection{Knowledge Graph Neural Networks}
To enable more effective learning on graph-structured data, researchers have dedicated to employing external knowledge\cite{DBLP:conf/www/BastosN0MSHK21, DBLP:conf/www/JinHL021}, such as large-scale knowledge bases Wikipedia and Freebase, to enrich node representations and apply them to downstream tasks. 
For example, 
KGAT \cite{DBLP:conf/kdd/Wang00LC19} presents a knowledge-aware recommendation approach
by explicitly modeling the high-order connectivity with semantic relations 
in collaborative knowledge graph.
COMPGCN \cite{DBLP:conf/iclr/VashishthSNT20} learns both node and relation embeddings in a multi-relational graph via using entity-relation composition operations from knowledge graph embedding.
Caps-GNN \cite{DBLP:conf/cikm/LiLZHWYW20} designs a novel personalized review generation approach with structural knowledge graph data and capsule graph neural networks.
Later on, RECON \cite{DBLP:conf/www/BastosN0MSHK21} points out that knowledge graph can provide valuable additional signals for short sentences, and develops a sentence relation extraction integrating external knowledge. 
KCGN \cite{DBLP:conf/aaai/HuangXXDXLBXLY21} proposes an end-to-end model that jointly injects knowledge-aware user- and item-wise dependent structures for social recommendation.
However, how to utilize the guidance of external knowledge to facilitate the understanding of node semantics within SIoT, and  further assess trustworthiness  among users is still an area that needs to be explored urgently. 

\section{Conclusion}
\label{section6}
In this paper, we present a novel knowledge enhanced graph neural network, namely KGTrust, for trust evaluation in Social Internet of Things.
In specific, we  comprehensively incorporate the rich node semantics within SIoT by deeply mining and encoding node-related information.
Considering that the
observed trust relationships are often relatively sparse, we use personalized PageRank-based neighbor sampling strategy to
enrich the trust structure.
To further maintain the multi-aspect properties of SIoT trust, we learn effective node embeddings
by employing a discriminative convolutional mechanism that considers the propagative and composable nature from the perspective of a user as a trustor or a trustee, respectively. 
After that, a learnable gating mechanism is introduced to adaptively integrate the information from dual roles of a user.
Finally, the learned embeddings for pairwise users are concatenated for a trust relationship predictor.
Extensive experimental results demonstrate the superior performance of the proposed new approach over  state-of-the-arts across three benchmark datasets.

% \begin{acks}
% This work is supported in part by the National Key Research and Development Program of China under Grant Nos. 2019YFB2102404, and the National Natural Science Foundation of China under grants 62272340.
% \end{acks}

\bibliographystyle{ACM-Reference-Format}
\bibliography{sample-base}

\newpage
\appendix

% \section{SUPPLEMENTARY MATERIAL}
% we first provide detailed poofs of import theorems in our paper i.e., Theorem 4.1, Proposition 4.2 and Theorem 4.3.
% Next, more experimental details are represented for reproduction.
\section{Notations}
Here we list the key notations in the main text in
Table 5.
\begin{table}[h!]
\setlength{\abovecaptionskip}{0.2cm}
\vspace{-0.2cm}
% \begin{center}
% \begin{minipage}{300pt}
	 \caption{\centering \label{tab:notations} Summary of notations.}
	\scalebox{0.99}{
	\begin{tabular}{p{1.1cm} p{6.5cm}}
		\toprule
		{\bf Notation}      & {\bf Description}\\
		\midrule
		$G$ & A Social Trust Internet of Things. \\
		$V, E$ & The sets of nodes and edges of a SIoT. \\
		$\bold{A}, \bold{D}$ & The adjacency matrix and node degree matrix.\\
	   % $\bold{D}$ & The node degree matrix. \\
	    $t_{ij}$ &  The trustworthiness from node $v_i$ to node $v_j$. \\
	   $\mathcal{N}_i$ & The set of neighbors of node $v_i$.\\
	    $(h, r, t)$ & A triplet in the knowledge graph. \\
	   $\bold{W}_{\psi}$ & The type-specific linear transformation matrix. \\
	    $\bold{P}$ & The personalized PageRank matrix. \\
        $\bold{h}_i, \bold{\overline{h}}_i$ & The latent embeddings of a given user node $v_i$ as trustor role or trustee role, respectively.\\
        $\sigma$ & The non-linear activation  function.\\
        $\alpha, \beta$ & Weights of type-level and node-level attentions.\\
		\bottomrule
	\end{tabular}
	}
	\label{notations}
% \end{minipage}
% \end{center}
\vspace{-0.2cm}
\end{table}

\section{Details of Experimental Settings}
\subsection{Datasets}
\begin{itemize}
\item \textbf{FilmTrust} is extracted from the film review website, which consists of two kinds of information, that is, social trust relationships between users; as well as interactive connections between users and objects.
% ratings that reflect the users' preference for objects, ranging from 0.5 to 4. 
For FilmTrust, considering that there is no information about users' comment behaviors and object names, we utilize random vectors to initialize user and object embeddings, which are trainable during the message passing process.

% and treat them as part of the learnable parameters of neural networks.

\item \textbf{Ciao} and \textbf{Epinions} are two who-trusts-whom knowledge-sharing websites that contain four types of information, that is, explicit trust 
relationships
between users; connections between users and objects representing interactive relationships; comment behaviors that reveal users' attitudes and preferences in the form of text; as well as object descriptions that reflect the characteristics of objects.
For these two datasets, we reserve users with more than 15 comment behaviors and objects with more than 10 comment behaviors.
\end{itemize}

\subsection{Baselines}
\begin{itemize}
 \item \textbf{GAT} \cite{DBLP:conf/iclr/VelickovicCCRLB18} is an attention-based graph neural network which assigns different weights to neighbors to improve the aggregating process.

\item \textbf{SGC} \cite{DBLP:conf/icml/WuSZFYW19} is a simplified graph neural network that reduces the complexity of model by removing nonlinearities and collapsing weight matrices between consecutive layers.

\item \textbf{SLF} \cite{DBLP:conf/kdd/XuH0D19} is a signed network embedding model, which
associates each type of social relationship to the comprehensive effects of positive and negative signed latent factors, and learns node embeddings by minimizing a negative log-likelihood objective function.

\item \textbf{STNE} \cite{DBLP:conf/icdm/XuHW0DY19} is a social trust network embedding model that preserves
both a node’s relations to latent factors and the trust transfer patterns for trust prediction.

\item \textbf{SNEA} \cite{DBLP:conf/aaai/LiTZC20} is a signed network embedding model through designing a graph attentional layer that utilizes a masked self-attention mechanism to calculate the importance coefficients of neighbors.

\item \textbf{DeepTrust} \cite{DBLP:conf/icdm/WangZYWHX19} is a trust evaluation model based on the homophily theory, which effectively combines the users' comment behaviors and the characteristics of users' interested objects for assessing trustworthiness. 

\item \textbf{AtNE-Trust} \cite{DBLP:conf/icdm/WangZ00ZX20} is a deep trust prediction model that captures user embeddings through taking into account of both the dual roles (trustor or trustee) and the connectivity properties of users.

\item \textbf{Guardian} \cite{DBLP:conf/infocom/LinGL20} is an end-to-end trust evaluation model that explicitly incorporates the popularity trust and engagement trust into user modeling using graph neural networks.
\end{itemize}

\section{Additional Experiments}
Following the setting from the
Section 4.4, we give the additional analysis of top $k$ for PPR-based neighbor sampling and the dimension of final embedding on FilmTrust dataset in Figure 5. As shown, 
the changing process of these two parameters follow the same way with the results in Figure 3 and Figure 4. 
\begin{figure}[htp]
\setlength{\abovecaptionskip}{0.2cm}
\centering
\subfigure[The number of top $k$]{
% \begin{minipage}[t]{0.2\linewidth}
\includegraphics[width=0.475\linewidth]{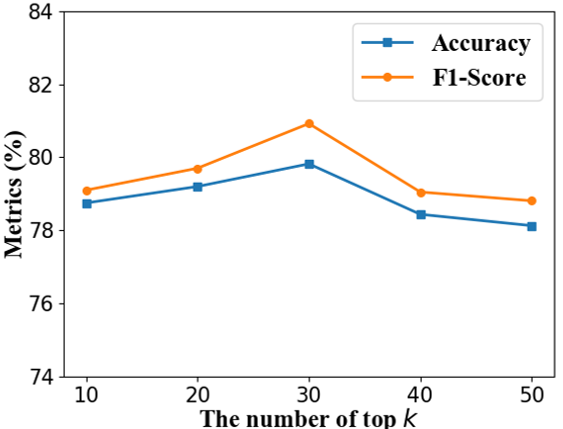}
% \end{minipage}
}
\subfigure[The dimension of final embedding]{
% \begin{minipage}[t]{0.2\linewidth}
\includegraphics[width=0.475\linewidth]{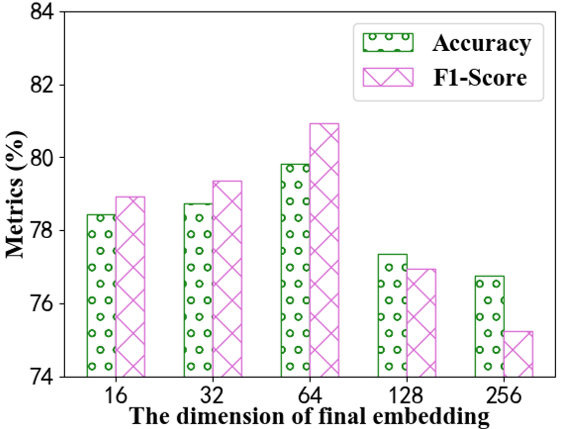}
% \end{minipage}
}
\caption{
Impact of hyper-parameter scope.}
\vspace{-0.25cm}
\end{figure}
\end{document}